\documentclass{prismshadow}
\usepackage{graphicx} 
\usepackage{xcolor} 
\usepackage{hyperref}
\usepackage{amsmath}
\usepackage{amssymb}
\usepackage{microtype}

\usepackage{natbib}
\usepackage{appendix}
\usepackage{booktabs}
\usepackage{url}
\usepackage{tcolorbox}
\usepackage{cleveref}
\usepackage{tabularx}
\usepackage{comment}
\usepackage{subfig}
\usepackage{tikz}
\usepackage[frozencache,cachedir=minted-cache]{minted}
\setminted{fontsize=\scriptsize}
\usepackage{amssymb}
\usepackage{wrapfig}

\usepackage{fontawesome}
\usepackage{pifont}

\usepackage{multirow}
\usepackage{makecell}
\usepackage{float}
\usepackage{xspace}

\newcommand{\bench}{\textit{GDPevo}\xspace}
\newcommand{\base}{\texttt{base}\xspace}
\newcommand{\fewshot}{\texttt{fewshot}\xspace}
\newcommand{\reflect}{\texttt{reflect}\xspace}
\newcommand{\self}{\texttt{self}\xspace}

\title{GDPevo: Evaluating Agent Self-Evolution on Real Business Tasks}

\author[1]{Leijun Zhou}
\author[1]{Zhihao Liu}
\author[1]{Xiang Qu}
\author[1]{Chenxu Liu}
\author[1]{Yifei Liu}
\author[1]{Yanke Yu}
\author[1]{Jingzhe Xu}
\author[1]{Xuejun Wu}
\author[1]{Buyue Qian}
\author[2]{Xi Chen}
\author[1]{Yaowei Zheng}
\author[1]{Junhao Hu}

\affiliation[1]{PrismShadow}
\affiliation[2]{New York University}

\abstract{
Agent self-evolution updates an agent's persistent state from prior experience and reuses it to solve related tasks more effectively. Evaluating self-evolution is difficult: existing benchmarks provide limited coverage of economically valuable task domains; do not always design training and test tasks such that test-time gains can be attributed to training experience; and remain vulnerable to data contamination. We present \bench, an evolution-native benchmark grounded in GDP-related enterprise workflows, together with the fully automated data pipeline that generates it. Its core mechanism, \textbf{rule hybridization}, decomposes each enterprise workflow into atomic business rules, distributes subsets of these rules across training tasks, and recombines them in held-out test tasks, so that test-time gains are attributable. \bench spans CRM, ERP, finance, healthcare, legal, and data-centric workflows. Its V1 release contains 120 tasks in 12 groups, with 5 training and 5 held-out test tasks per group; full automation lets the pipeline expand the suite to 240 tasks in 24 groups (V2) within two days, providing a practical response to contamination. Using \bench, we evaluate four agents (harness + model) under four supervision types. Self-evolution consistently improves held-out accuracy by up to $16.44$~pp. But the best evolved agents remain far below a fully informed oracle ceiling of $91.6\%$, indicating that the self-evolution ability of current agents remains far from fully realized. We publicly release the pipeline, benchmark, and full evaluation results: \url{https://github.com/Prism-Shadow/GDPevo}.
}

\date{\today}
\correspondence{Junhao Hu (\email{hu@prismshadow.com}), Yaowei Zheng (\email{zheng@prismshadow.com})}

\begin{document}

\maketitle

\section{Introduction}
\label{sec-intro}

In the AI era, once a task can be \emph{automated} and \emph{evaluated}, it is rarely far from being solved at scale. To bring these two capabilities to agentic task solving, agent self-evolution has emerged as a promising paradigm. Under this paradigm, an agent continuously improves task performance by updating its internal state, such as parametric states and non-parametric states (e.g., memory, skills, prompts, or harness code), based on prior interactions and reusing the resulting state in subsequent related tasks~\citep{gao2026survey}. Automation includes techniques, such as continual learning~\citep{wang2024cl}, experiential learning~\citep{shinn2023reflexion,zhao2024expel}, and recursive self-improvement~\citep{schmidhuber2007godel,zhang2025dgm}, for performing these internal updates. Evaluation, on the other hand, determines whether these updates actually improve subsequent behavior. Although the literature now offers many techniques to automate self-evolution, evaluation methodology remains less mature.

We focus on evolution benchmarks that measure learning through an explicit train--test split\footnote{A separate family of benchmarks instead measures evolution over time, tracking how performance improves with accumulated experience within a single long-horizon deployment (e.g., EdgeBench~\citep{zhu2026edgebench}) rather than across a train--test split.}, falling into two categories. First, evolution-native benchmarks are designed to test whether an agent learns from training experience and transfers what it learns to held-out tasks. They deliberately construct the train--test split so that the capabilities an agent acquires on the training tasks are genuinely required and exercised on the test tasks~\citep{zhang2026skillflow,jiang2026seaeval,gao2026evoagentbench,huang2026benchtrace}. Second, evolution-adaptive benchmarks repurpose task suites originally designed for static, single-episode capability measurement, such as interactive reasoning environments and software-engineering tasks~\citep{shinn2023reflexion,zhao2024expel,jimenez2024swebench,merrill2026terminalbench}. They introduce no deliberate design and directly split the suite into training and test sets, reporting whether the experience gained while solving the training set improves performance on the test set.

Existing benchmarks leave three limitations. First, evolution-native benchmarks provide limited coverage of difficult, economically valuable tasks found in domains such as finance, law, and healthcare~\citep{patwardhan2025gdpval,nandi2025sopbench,li2026jobbench}. Tasks such as invoice auditing, compliance checks, and record reconciliation are governed by business-specific rules that an agent must learn and follow, and their correctness can be verified against deterministic criteria, making them well suited for studying evolution. Second, evolution-adaptive benchmarks provide broad and difficult tasks, but because their inherited splits of train--test sets were never aligned to any notion of transferable ability, an accuracy gain after evolution is not attributable. Even evolution-native benchmarks often begin with existing benchmarks and search for plausible relationships from which to form training and test splits~\citep{zhang2026skillflow,gao2026evoagentbench,huang2026benchtrace}. Because such relationships are extracted post hoc from existing benchmarks, they are inherently limited in number and diversity. Third, both categories of benchmarks release static and public task sets, leaving them exposed to data contamination~\citep{white2025livebench,wu2025antileakbench,chen2025contamination}. In an era of rapidly advancing AI, any benchmark without an explicit mechanism to counter contamination quickly loses its validity.

\bench addresses these limitations as an evolution-native benchmark. First, it is the first benchmark to evaluate self-evolution on GDP-related tasks, spanning CRM, ERP, finance, healthcare, legal, and data-centric domains. Second, we introduce \textbf{rule hybridization} to design the train--test relationship from the outset. The construction pipeline decomposes each domain workflow into atomic business rules, plants subsets of these rules across five training tasks, and recombines them across five held-out test tasks. An agent must infer reusable rules from the training sets and apply them compositionally at test time, so that test-time gains are attributable. Third, \bench uses an automatic construction pipeline (Figure~\ref{fig-pipeline}) to expand the benchmark as public tasks become exposed. This automation lowers the marginal effort of generating new task groups and provides a practical response to contamination. \bench is therefore both a construction pipeline and the benchmark generated by that pipeline.

The first version (V1) of \bench contains 120 tasks in 12 task groups, with 5 training and 5 held-out test tasks per group (Table~\ref{tbl-data}). Thanks to full automation, we scale the benchmark to 240 tasks in 24 groups (V2) in just two days. The benchmark has three additional properties. First, every task uses a deterministic, rule-based grader instead of relying on an LLM as a judge. An LLM transforms independent rubric points into code-based test cases, which make scores reproducible and trace each failure to a violated rule. Second, \bench treats cost as a first-class metric. Tokens, agent turns, and monetary cost accompany accuracy because a useful evolution strategy should improve both task success and resource efficiency. Third, its reports provide per-group breakdowns, radar views, and transfer heatmaps, so users can trace aggregate gains to specific domains and examine how evolution on a source domain affects a target domain, since not all such transfer is beneficial.

Unless otherwise specified, all experiments in this paper use the combined V1 and V2 data. Evolution has two dimensions: the \emph{supervision type} determines what signal the agent may draw on during training, and the \emph{evolution method} determines how that signal becomes persistent state. We evaluate four agents (harness + model) under four supervision types: a no-evolution \base; \fewshot, which uses the training questions and their gold answers, akin to supervised fine-tuning; \reflect, which uses the training questions and the scores of its own attempts rather than the gold answers, akin to reinforcement learning; and \self, which uses only the training questions, akin to unsupervised learning. Throughout, the evolution method is skill-based. Our study is organized around three research questions (Section~\ref{sec-eval}). \textbf{RQ1}: how much does each agent benefit from each supervision type, in terms of both accuracy and cost? We find that \fewshot is the most reliable supervision type, that every evolved combination improves over its \base (by up to $16.44$~pp) and can even reduce test-time cost, yet all agents remain far below a fully informed oracle ceiling of $91.6\%$. \textbf{RQ2}: how well do evolved skills transfer across task groups of different domains? We find that \fewshot, like supervised fine-tuning, overfits its source group and can hurt on others, whereas \reflect, like reinforcement learning, transfers more robustly. \textbf{RQ3}: which matters more for evolution, the evolution method inside the harness, such as the skill creator, or the model itself? We find that even a minimal evolution method already works well, so the magnitude of evolution comes mainly from the model's intelligence.

The main contributions of this paper are as follows:
\begin{itemize}
    \item We propose \textbf{rule hybridization}, which makes generalization from training to test concrete and test-time gains attributable.
    \item We conduct a comprehensive evaluation of four agents (harness + model) under four supervision types.
    \item We release the fully automated \bench generation pipeline and its generated benchmark (both the V1 and V2 versions).
\end{itemize}
\section{Background and Related Work}
\label{sec-related}

This section positions \bench by reviewing which internal states agent self-evolution automatically updates and how these updates are evaluated.

\subsection{Self-Evolving Agents}

An agent's behavior can be viewed as sampling from \(P(y \mid C, \theta)\), where \(y\) denotes the output, \(\theta\) denotes the model parameters, and \(C\) denotes the context that conditions its behavior. Fine-tuning and reinforcement learning update \(\theta\) to shift the output distribution toward desired behavior; when such updates continue across tasks, they represent parametric continual learning~\citep{wang2024cl}. By contrast, prompt, context, and harness engineering modifies \(C\) to constrain the output distribution and increase the likelihood of task-specific responses. Here, \(C\) includes the prompt and persistent components such as memories, skills, and even the harness code. For example, Reflexion and ExpeL store feedback or reusable lessons in memory~\citep{shinn2023reflexion,zhao2024expel}, while Voyager and SkillFlow build and maintain executable skill libraries~\citep{wang2024voyager,zhang2026skillflow}. Other techniques update prompts or harness code. For example, GEPA evolves prompts through reflective feedback~\citep{agrawal2025gepa}, and the Darwin G\"odel Machine edits agent code and retains variants that improve benchmark performance~\citep{zhang2025dgm}.

In this paper, our experiments focus on a non-parametric evolution method in the form of skills. An agent uses a skill creator to learn skills from the training tasks and then carries the resulting skill library into the test tasks. This choice defines our experimental setup rather than the scope of the benchmark: \bench is agnostic to the form of persistent state being updated. It can evaluate evolution through memories, prompts, skills, harness code, or model parameters.

\subsection{Evaluating Self-Evolution}

Evolution benchmarks evaluate whether an agent can learn from prior experience and transfer what it learns to unseen, related tasks, and they fall into two categories. The first category does not define an explicit train--test split: EdgeBench~\citep{zhu2026edgebench} fits scaling laws to an agent's learning curve over ultra-long-horizon deployment within a single environment, whereas RSIBench~\citep{meng2026rsibench} evaluates the data-centric research loop that recursively improves a fixed target model. The second category explicitly defines a train--test split, which is the setting this paper focuses on. Within this setting, we review two types of benchmarks defined by how they construct the train--test relationship---\textbf{evolution-native} and \textbf{evolution-adaptive}---and then discuss the motivations for a new benchmark.

\subsubsection{Evolution-Native Benchmarks}

The proper evaluation of self-evolution requires a carefully designed train--test relationship. If the training and test tasks are identical or nearly identical, improvements may reflect memorization rather than generalization. If they are unrelated, experience gained during training may provide no relevant support for test performance. These benchmarks must therefore use distinct but related tasks and deliberately construct the train--test split so that capabilities acquired during training are both required and exercised on held-out test tasks.

For example, EvoAgentBench~\citep{gao2026evoagentbench} constructs ability graphs and a 528/267 train--test split in which every test task has verified procedural support from the training tasks. SkillFlow~\citep{zhang2026skillflow} organizes 166 tasks into 20 workflow families to evaluate skill discovery, repair, and reuse across sequential tasks. SEA-Eval~\citep{jiang2026seaeval} measures both task success and token consumption along sequential tasks, capturing the effectiveness and efficiency of evolution.

However, existing evolution-native benchmarks provide limited coverage of more difficult and economically valuable enterprise workflows. Moreover, these benchmarks derive train--test relationships post hoc from existing benchmarks, limiting the diversity of transferable knowledge or procedures. In contrast, \bench targets difficult, economically valuable enterprise workflows and constructs train--test relationships from the ground up, using diverse business rules, enabling a broader range of knowledge and procedural transfer.

\subsubsection{Evolution-Adaptive Benchmarks}

Unlike evolution-native benchmarks, evolution-adaptive benchmarks do not explicitly construct train--test pairs around transferable abilities. Because their direct train--test splits are not aligned with any notion of transfer, performance gains after evolution cannot be reliably attributed to learning from prior experience. Nevertheless, by reusing a wide range of established task suites, these benchmarks offer greater task diversity and broader coverage of difficult problems.

For example, Reflexion~\citep{shinn2023reflexion} and ExpeL~\citep{zhao2024expel} evaluate learning on repeated or partitioned tasks from ALFWorld, WebShop, HotpotQA, HumanEval, and related environments. Coding-agent systems commonly use SWE-bench~\citep{jimenez2024swebench}, which contains 2,294 repository issues paired with executable tests, or realistic command-line suites such as Terminal-Bench~\citep{merrill2026terminalbench}. The Darwin G\"odel Machine~\citep{zhang2025dgm}, for instance, selects self-modified coding agents based on their benchmark performance.

\bench is an evolution-native benchmark whose train--test relationships are explicitly constructed to evaluate transfer from experience.

\subsection{Further Requirements for Evolution Benchmarks}

Beyond economically valuable workflows and train--test splits designed around transferable rules, evolution benchmarks require four additional properties. First, they should resist data contamination through a pipeline that can refresh exposed tasks~\citep{white2025livebench,wu2025antileakbench,chen2025contamination}. Second, they should use deterministic, rule-based graders rather than LLM judges, so that scores are reproducible and each error can be traced to a specific rule violation. Third, they should treat tokens, agent turns, and monetary cost as first-class metrics alongside accuracy, because a good self-evolving agent should not only become more accurate but also more efficient. Finally, they should provide diagnostic views, such as score breakdowns and transfer heatmaps, that reveal where and how evolution succeeds. Together with the two requirements identified in the preceding subsections, these considerations motivate the design of \bench, which we introduce in the next section.

\begin{figure}[t]
\begin{center}
\centerline{\includegraphics[width=0.92\textwidth]{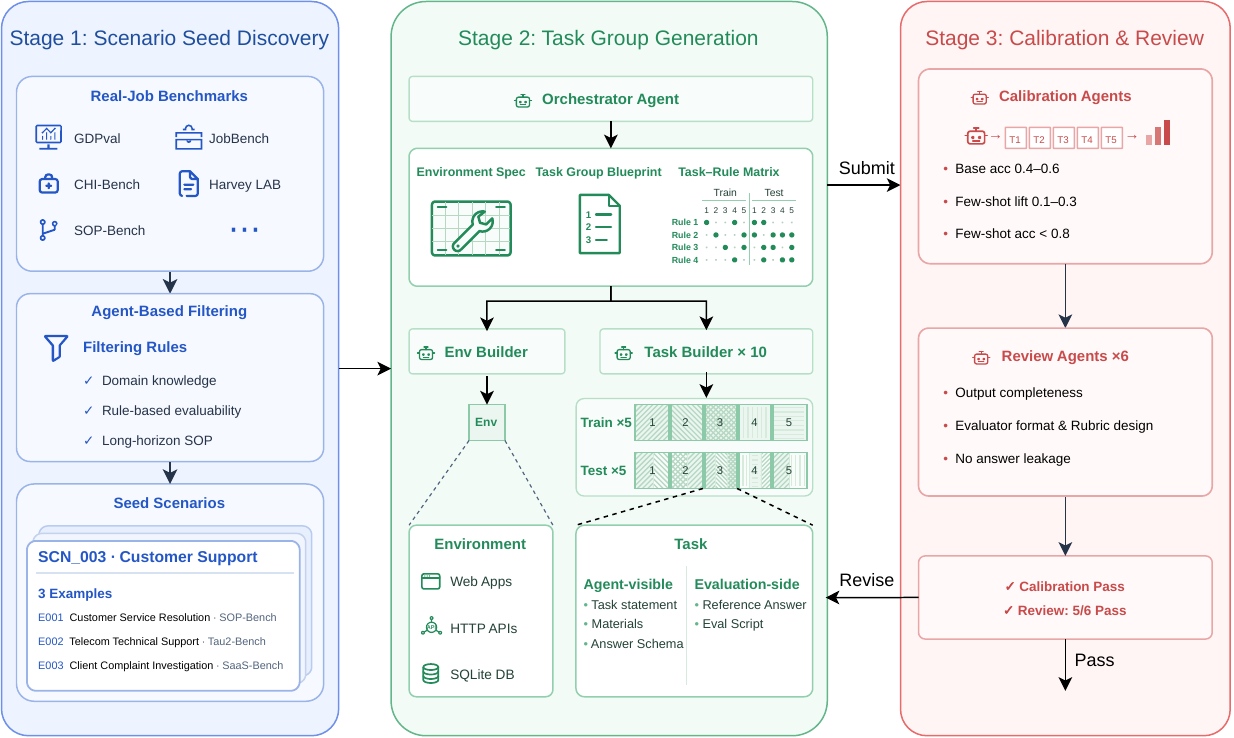}}
\caption{The \bench data construction pipeline. Every agent in this pipeline uses the same harness and model, Codex with GPT-5.5, but each runs as a separate instance with its own context, so the agents remain relatively independent.}
\label{fig-pipeline}
\end{center}
\vskip -0.2in
\end{figure}

\section{The \bench Benchmark}
\label{sec-bench}

The unit of \bench is a \textbf{task group}: one shared business environment (e.g., a medical-records system or an employee database), 5 training tasks, and 5 held-out test tasks that use the same environment. An agent processes the 5 training tasks, optionally with supervisory signals indicating whether its attempts are correct, and updates its persistent state accordingly; it then solves the held-out test tasks. The accuracy gain over the same agent given no training experience measures improvement after evolution. This section describes how the \bench pipeline constructs the benchmark; Section~\ref{sec-eval} describes how we evaluate self-evolution with the benchmark.

\subsection{Fully Automated Construction Pipeline}
\label{sec-bench-pipeline}

\bench is constructed end to end by agents in three stages, as shown in Figure~\ref{fig-pipeline}; the only human involvement is collecting the domain benchmarks that kick off the process. First, \textbf{seed scenario discovery} takes these existing domain-specific benchmarks as input, such as SOP-Bench~\citep{nandi2025sopbench} and GDPval~\citep{patwardhan2025gdpval}, and an agent proposes candidate scenarios. Second, \textbf{task group generation} takes each scenario and its associated examples and produces a task group: one shared environment (e.g., a medical-records system or an employee database) together with 10 related tasks, split into 5 training and 5 held-out test tasks. Third, \textbf{calibration and review} checks each task group for quality, difficulty, and diversity, keeping the groups that pass and revising the rest until they do.

Because this pipeline is fully automated and agent-generated, \bench can counter data contamination by rapidly regenerating a fresh version whenever a released one risks exposure. First, our initial release, V1, contains 120 tasks in 12 task groups, each following the 5+5 training/test structure: 4 groups in CRM, 4 in ERP, and 4 in finance (Table~\ref{tbl-data}). Second, in under two days the same pipeline produced V2, adding another 120 tasks in 12 task groups with the same 5+5 structure, spanning 4 groups in healthcare, 4 in legal, and 4 in data-centric work. V1 and V2 together form a single benchmark of 240 tasks in 24 groups; unless stated otherwise, all experiments in this paper use the full 240-task benchmark.

\subsection{Selecting economically valuable seed scenarios}
\label{sec-bench-seeds}

To construct difficult and diverse tasks over economically valuable work, we draw from public benchmarks of real work, including GDPval~\citep{patwardhan2025gdpval}, SOP-Bench~\citep{nandi2025sopbench}, and JobBench~\citep{li2026jobbench}, and an agent expands these candidate benchmarks into concrete seed scenarios and their associated examples, following three criteria. First, the scenario must embed domain-relevant hidden rules; rule hybridization later recombines these rules across tasks (Section~\ref{sec-bench-hybrid}) so that an agent must genuinely learn them from the training tasks and generalize to the held-out test tasks. Second, the rules must be measurable, complete, and deterministic, so that task outcomes can be checked by a deterministic rule-based grader (Section~\ref{sec-bench-grading}). Third, the scenario must be complex enough to sustain a long-horizon standard operating procedure (SOP), and the scenarios must be diverse.

For example, Figure~\ref{fig-pipeline} shows the seed scenario \textbf{service ticket resolution}. It asks an agent to verify a customer's account, diagnose the reported service fault, carry out a repair, and record its final disposition. This workflow carries many enterprise-specific rules: which account states are eligible for support, which diagnostic readings count as a fault, and which faults must be escalated and to which team. Every such rule resolves to a checkable outcome, so a rule-based grader can score it. Each scenario becomes one task group of 10 tasks, and those tasks draw their concrete problem statements, rules, and reference answers from \textbf{examples}: specific instances of the scenario that we collect alongside it. For more detailed examples of seed scenarios, see Appendix~\ref{sec-appendix-tasks}.

\subsection{Separating training and test through rule hybridization}
\label{sec-bench-hybrid}

To make agents genuinely learn transferable rules from the training tasks and apply them to the held-out test tasks, we introduce \textbf{rule hybridization}. First, for each scenario (which becomes one task group), the pipeline decomposes the business logic of its examples into \textbf{atomic rules}, which are minimal and independently checkable decision rules. These rules are deliberately absent from the model's world knowledge; they are internal conventions specific to a particular enterprise, such as one company's sponsor-status priority, another company's blacklist exclusion, and a company-specific invoice expiration date. Second, the pipeline scatters the rules across the 5 training tasks, so each training task exposes only a subset of these rules. An agent attempts these training tasks; because the hidden rules do not exist in world knowledge, it cannot know them at first and will most likely fail. During evolution, we supply various supervisory signals that let the agent infer these rules and record them in its skills. Third, the pipeline recombines the rules across the 5 test tasks: a test task may invoke the priority and blacklist rules together even though no training task contains that combination. Only an agent that has learned the rules during training and can compose and apply them solves the test tasks.

Stage 2 of Figure~\ref{fig-pipeline} implements this construction in two steps. First, an orchestrator agent turns a scenario and its examples into a blueprint for the task group: an environment specification, the 10 task statements, the atomic rules extracted from the examples, and a task--rule matrix that fixes which rules each training and test task exposes. Second, the orchestrator spawns one environment builder to realize the shared environment and 10 task builders that each construct a single task from its assigned rule subset.

\subsection{Calibration and independent review}
\label{sec-bench-quality}

We ask a calibration agent to attempt each candidate task group under three requirements. First, solving the 5 test tasks directly, without any training experience, should yield a score of roughly 40--60\%; this band rejects task groups that are already solvable from the model's world knowledge alone as well as those that are too difficult. Second, when the agent first learns from the 5 training tasks and evolves before solving the test tasks---here we use \fewshot, the supervision type that currently performs best (Section~\ref{sec-eval})---the test score must improve by about 0.1--0.3; this rejects task groups whose hidden rules either cannot be learned from training or are too easy to learn. Third, even after evolution the final score must stay below 0.8, so that measurable headroom remains.

Finally, six independent reviewer agents inspect each candidate task group separately, and a group is accepted only if at least 5 of the 6 vote to accept it. They check three aspects. First, output completeness: builder agents may declare completion while leaving files, evaluators, or environments incomplete owing to model laziness~\citep{liu2025llmigrate}, so reviewers verify that every required artifact is present. Second, evaluation format and rubric design: the rubrics should be well-formed and not overly similar to one another, presenting a degree of diversity that the reviewers agree on; because diversity is hard to quantify, we rely on the reviewer agents to judge it. Third, no answer leakage: when organizing a task, its answers must reside in a standalone, separate folder and must not appear anywhere else.
\section{Evaluation}
\label{sec-eval}

\subsection{Evaluation Setup}
\label{sec-eval-setup}

\paragraph{Dataset.}
All experiments use the full \bench benchmark of 240 tasks in 24 task groups, combining V1 (CRM, ERP, and Finance) and V2 (healthcare, legal, and data-centric work). Each task group shares one business environment and contains 5 training tasks and 5 test tasks. The training tasks expose fragments of the group's hidden business rules, whereas the test tasks recombine those rules.

\paragraph{Metrics.}
\label{sec-bench-grading}
To make the evaluation deterministic, comprehensive, and interpretable, \bench provides three complementary forms of measurement. First, every task is scored by a deterministic, rule-based grader rather than an LLM judge. Each task comes with a set of weighted rubric points, but instead of asking an LLM to decide whether each point is satisfied, we ask an LLM to translate each rubric point into a code-based test case, which makes grading deterministic across runs. Each point earns either its full assigned weight or zero, and a task's score is the normalized sum of its passed point weights. Every test task is run three times, and all reported accuracies are means over those three runs. Second, \bench reports costs alongside accuracy. Each run records token consumption, agent turns, and monetary cost derived from model prices. Cost is a first-class metric because a good self-evolving agent should be not only accurate but also efficient. Third, \bench provides multiple views of the same results. Aggregate tables and radar maps compare different agents and supervision types; per-task score breakdowns reveal which rubric points are satisfied and, in turn, which hidden rules the agent discovered and which it missed; and transfer heatmaps expose which training experiences help---or hurt---which held-out groups.

\paragraph{Agents.}
We define an agent as the composition
\begin{equation}
    \text{Agent} = \text{Harness} + \text{Model}.
\end{equation}
The harness determines tool use, context management, skill loading, and execution control, while the model supplies the underlying reasoning policy. We evaluate two harnesses, Codex and Claude Code.\footnote{Codex: \url{https://openai.com/codex/}; Claude Code: \url{https://docs.anthropic.com/en/docs/claude-code/overview}.} Candidate models include GPT-5.5, Opus-4.8, GLM-5.2, and DeepSeek-V4-Pro-Preview. The complete set of harness--model combinations, together with their versions, thinking levels, and full results, is provided in Appendix~\ref{sec-appendix-full-results}.

\paragraph{Supervision types.}
Evolution has two independent dimensions. The \emph{supervision type} determines what signal the agent may draw on while working through the training tasks; the \emph{evolution method} determines how that signal is turned into a skill. This paragraph lists the four supervision types, and the next lists the evolution method.

We compare four supervision types. First, in \base, the agent receives no training experience and directly attempts the test tasks. Second, in \fewshot, the agent is given the questions, environment, and gold answers of the 5 training tasks; it reflects on how those gold answers are derived and distills the resulting experience into a reusable skill, then attempts the test tasks with that skill. This supervision is closely analogous to supervised fine-tuning (SFT): both learn from input--output supervision, except that \fewshot updates a textual skill rather than model weights~\citep{brown2020fewshot}. Third, in \reflect, the agent receives the questions and environment of the 5 training tasks but not their gold answers, and attempts to solve them on its own; we then run our rule-based grader and return the score as feedback, from which the agent learns and revises its experience and strategy. This process repeats three rounds (\reflect-3) and is closely analogous to reinforcement learning (RL)~\citep{shinn2023reflexion,song2025icrl}. Unless noted otherwise, \reflect refers to this three-round setting. Fourth, in \self, the agent receives only the questions and environment of the 5 training tasks and attempts them with no answer-related supervision. It nonetheless has the chance to explore the environment: what the database contains, which fields hint at the underlying rules, and, at the least, how to query the database and operate the environment efficiently, so that at test time it can skip this exploration. This supervision is similar to unsupervised learning purely from text rather than from parameters.

\paragraph{Evolution method.}
The evolution method determines how the available supervision is turned into persistent state. Self-evolution can update many kinds of persistent state, from model parameters to context, and, within context, memory, skills, or even the agent harness. \bench is agnostic to this choice and can evaluate any of these forms (Section~\ref{sec-related}). All experiments in this paper adopt a skill-based evolution method, following recent evolution-native benchmarks such as EvoAgentBench~\citep{gao2026evoagentbench} and SkillFlow~\citep{zhang2026skillflow}. Our evolution state is thus a portable \texttt{SKILL.md} artifact: for each task group, an agent attempts the 5 training tasks, consumes the supervision allowed by the current supervision type, and distills transferable business rules, environment-use procedures, output conventions, and common failure modes into the skill; a new agent then loads the skill and attempts the 5 test tasks.

Within a skill-based method, the \emph{skill creator} is the component that performs this distillation. In \textbf{RQ3}, we compare five skill creators while holding the agent fixed: (1) \textit{Naive}, a minimal baseline we write ourselves---a single sentence that asks the agent to summarize experience from the available information and record it as a skill---together with the built-in skill creators of (2) Claude Code, (3) Codex, (4) OpenCode, and (5) deepagents.

\paragraph{Agent-driven evaluation.}
Beyond launching experiments by hand or via scripts, we mostly drive the whole evaluation with another agent (typically Codex with GPT-5.5) that runs every experiment and renders every figure from a natural-language request.


\paragraph{Environment.}
Every training and testing attempt executes in an independent Docker container with a dedicated working directory and harness home. Containers mount only allowlisted files, can run in parallel, and cannot read other attempts, source answers, evaluator files, or prior traces. Each raw trace, generated skill, answer, score, and run metadata record is written beneath its own attempt directory, preserving isolation between training and test and across agents.

\subsection{RQ1: How Much Does Each Agent Benefit from Each Supervision Type?}
\label{sec-eval-main}

\begin{figure}[t]
\centering
\includegraphics[width=\columnwidth]{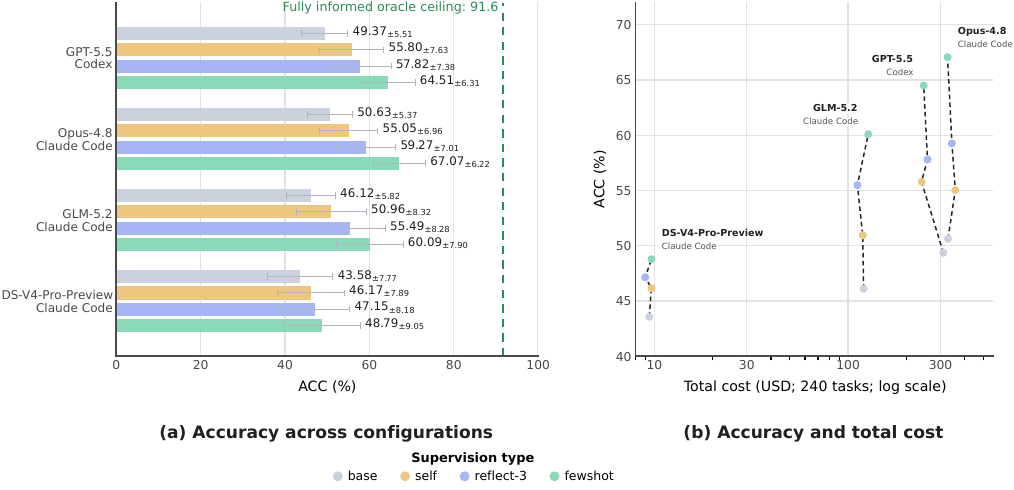}
\caption{Accuracy--cost trade-offs on \bench. (a) Mean accuracy across task groups for each model--harness configuration and supervision type; each test task is averaged over three runs, and error bars indicate one standard deviation across task groups. The green dashed line marks the fully informed oracle ceiling ($91.6\%$). (b) Accuracy and total evaluation cost over all 240 tasks for each model--harness--supervision configuration. Colors denote supervision types, and dashed lines connect configurations sharing the same model and harness. Skill-generation overhead is excluded and can be found in Table~\ref{tbl-evolution-cost}.}
\label{fig-eval-main}
\end{figure}

Figure~\ref{fig-eval-main} and Tables~\ref{tbl-full-leaderboard}--\ref{tbl-evolution-cost} yield five findings. First, \fewshot attains the highest accuracy for all four agents, and every agent surpasses its \base when provided with any form of supervision, by $2.59$ to $16.44$~pp. Second, evolution can substitute for model training. DeepSeek-V4-Pro-Preview \fewshot attains accuracy comparable to GPT-5.5 \base ($48.79\%$ versus $49.37\%$) at approximately $1/28$ of the amortized end-to-end cost, and GLM-5.2 \fewshot surpasses GPT-5.5 \base and Opus~4.8 \base by $10.72$ and $9.46$~pp, respectively, at roughly half the cost. Third, a weaker starting point does not entail greater evolution headroom: DeepSeek-V4-Pro-Preview has the lowest \base accuracy ($43.58\%$) yet the smallest \fewshot gain ($+5.21$~pp), whereas Opus-4.8 has the highest \base accuracy ($50.63\%$) and the largest gain ($+16.44$~pp).

Fourth, evolution does not only improve accuracy; it can also make test-time execution cheaper. GPT-5.5 \fewshot raises accuracy by $15.14$~pp while reducing test-time cost by $20.88\%$; amortizing its one-time skill-generation cost (Table~\ref{tbl-evolution-cost}) over the five held-out tasks yields $\$1.296$ per task, nearly identical to the $\$1.294$ \base cost. Opus~4.8, in contrast, achieves the largest accuracy gain with essentially unchanged test-time cost ($-0.57\%$), yet its amortized end-to-end cost rises from $\$1.37$ to $\$1.87$ per task. Fifth, evolution with clear supervision signals forms the Pareto frontier as shown in Figure~\ref{fig-eval-main}. The Pareto frontier consists of four \fewshot agents together with GLM-5.2 \reflect and DeepSeek-V4-Pro-Preview \reflect.

To interpret these gains against an upper bound, we estimate an oracle ceiling. Ideally, we would give human domain experts all of a task group's hidden rules and measure how well they solve the test tasks, which would reveal the headroom available to any supervision type or evolution method. Because we lack experts across all covered domains, such as legal, medical, and financial workflows, we approximate this ceiling with a fully informed model: we provide the model with all hidden rules together with the training questions and their gold answers, so that it need not learn or infer anything and simply applies the given rules to the test tasks. Under this full-information setting the model reaches $91.6\%$ (the dashed lines in Figure~\ref{fig-eval-main}; see Appendix~\ref{sec-appendix-human}). An agent that could genuinely learn all hidden rules from experience and apply them compositionally should therefore approach this level. Since the best evolved configurations remain well below it, the self-evolution ability of current agents is still limited, and how to make agents evolve more effectively warrants further study.

The appendix provides further detail behind these aggregate numbers. Appendix~\ref{sec-appendix-case-studies} traces individual rules from training examples to held-out tests across four contrasting cases: two in which an agent genuinely learns the hidden rules and applies them compositionally (Sections~\ref{subsec:appendix-good-case-024} and~\ref{subsec:appendix-positive-case-014}), one in which it learns only a narrow fragment of them and thus underfits (Section~\ref{subsec:appendix-limited-case-016}), and one in which it overfits its training experience by applying rules beyond their valid scope, ending up worse than \base (Section~\ref{subsec:appendix-overfit-case-018}); Section~\ref{subsec:appendix-cross-case-discussion} synthesizes what distinguishes a transferable rule from a memorized one. Appendix~\ref{sec-appendix-breakdowns} adds diagnostic views: rubric-level score breakdowns showing which individual rules an agent begins to satisfy after evolution (Figure~\ref{fig:rubric-breakdown}), and task-group radar charts showing where gains concentrate across domains (Figure~\ref{fig:task-group-radar}). The complete leaderboard and the one-time skill-generation costs are reported in Appendix~\ref{sec-appendix-full-results}.

\subsection{RQ2: How Well Do Evolved Skills Transfer Across Domains?}
\label{sec-eval-transfer}

\begin{figure}[t]
\centering
\includegraphics[width=0.92\columnwidth]{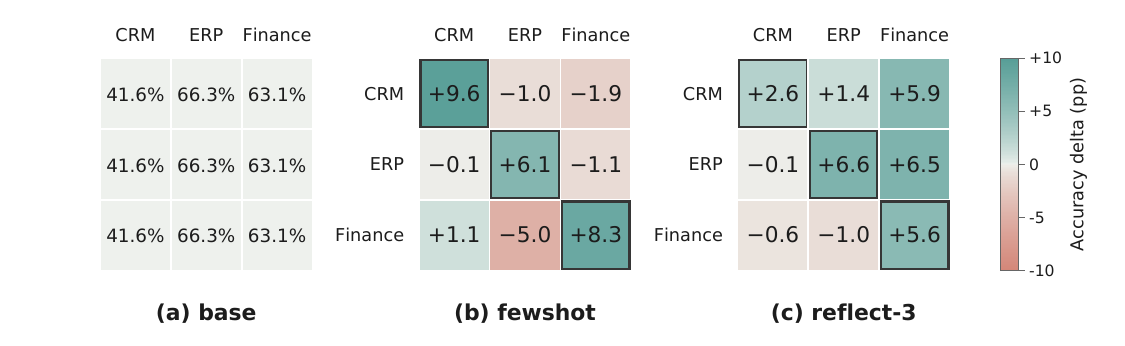}
\caption{Cross-domain evolution transfer across CRM, ERP, and Finance. (a) reports target-group \base accuracy. (b) and (c) show accuracy deltas using \fewshot and \texttt{reflect-3}, respectively, relative to the corresponding \base accuracy. Rows denote source groups used for training and columns denote target groups used for testing.}
\label{fig-eval-transfer}
\end{figure}

The previous subsection shows that, within a single task group, an agent that learns from the 5 training tasks acquires within-domain knowledge that helps it solve the 5 held-out test tasks. A natural next question is whether training on a \emph{different} domain still helps: if the 5 training tasks come from another domain, does the resulting experience (skills) benefit the target domain? To answer this question, we run a cross-domain transfer experiment. To keep it tractable, we restrict the experiment to the three V1 domains---CRM, ERP, and Finance---and randomly pick one task group from each, namely tg02 for CRM, tg06 for ERP, and tg10 for Finance; for the same reason we drop \self and compare only \fewshot and \reflect. We then cross these three groups: an agent trains on one group and is tested on all three, yielding the $3\times3$ source--target matrix in Figure~\ref{fig-eval-transfer}, where rows are the source (training) group and columns the target (test) group.

Figure~\ref{fig-eval-transfer} supports three findings. First, the diagonal cells---training and testing on the same group---are consistently positive, and under \fewshot they are the largest gain in every row, matching the within-domain results reported in Section~\ref{sec-eval-main}. Second, off the diagonal, \fewshot behaves much like SFT: it tends to overfit the source domain, so transferring its generated skill across domains mostly fails to help and is often harmful, with five of the six off-diagonal cells negative and a worst case of $-5.0$~pp (training on Finance, testing on ERP). Third, \reflect behaves much like RL, whose overfitting is milder: half of its off-diagonal cells are positive (up to $+6.5$~pp when transferring from ERP to Finance) and its worst case loses only $1.0$~pp, so cross-domain evolution rarely hurts and sometimes still helps. Under \reflect, however, the diagonal no longer dominates: evolving on CRM helps Finance ($+5.9$~pp) more than it helps CRM itself ($+2.6$~pp), and evolving on ERP helps Finance ($+6.5$~pp) nearly as much as ERP ($+6.6$~pp). On-policy feedback thus appears to yield more general skills, at the cost of some within-domain specialization.

\subsection{RQ3: Which Matters More, the Evolution Method or the Model?}
\label{sec-eval-methods}

\begin{table}[t]
\centering
\small
\setlength{\tabcolsep}{7pt}
\renewcommand{\arraystretch}{1.22}
\begin{tabular}{@{}lrrrr@{}}
\toprule
& \multicolumn{2}{c}{\textbf{GPT-5.5 / Codex}} & \multicolumn{2}{c}{\textbf{DS-V4-Pro-Preview / Codex}} \\
\cmidrule(lr){2-3}\cmidrule(lr){4-5}
\textbf{Evolution method} & \textbf{ACC (\%)} & \textbf{$\Delta$ (pp)} & \textbf{ACC (\%)} & \textbf{$\Delta$ (pp)} \\
\midrule
\base                 & 49.66 & --      & 42.48 & --     \\
CC creator            & 62.15 & $+12.49$ & 46.74 & $+4.26$ \\
Codex creator         & 62.19 & $+12.53$ & 47.05 & $+4.57$ \\
DeepAgents creator    & 62.69 & $+13.03$ & 47.75 & $+5.27$ \\
OpenCode creator      & 60.79 & $+11.13$ & 47.84 & $+5.36$ \\
\textbf{Naive creator} & \textbf{65.12} & $\mathbf{+15.46}$ & \textbf{48.01} & $\mathbf{+5.54}$ \\
\bottomrule
\end{tabular}
\caption{Controlled evolution-method comparison. We fix the agent and the supervision type (\fewshot) and vary only the evolution method, which here means the skill creator. Every test task is run three times; each reported ACC is the mean over those three runs, macro-averaged over the 24 task groups, and $\Delta$ is the lift over the same agent's \base.}
\label{tbl-eval-methods}
\end{table}

In the experiments so far, each agent evolves on its own: it starts from its own \base accuracy and improves to its own \fewshot accuracy. Here we instead fix a single agent---so there is a single \base accuracy---and fix the supervision type to \fewshot, leaving the evolution method as the only variable, to see how much the method affects the effectiveness of evolution. Since the method is skill-based throughout, varying it means varying the skill creator: we compare the five creators of Section~\ref{sec-eval-setup} on two agents, Codex with GPT-5.5 and Codex with DeepSeek-V4-Pro-Preview.

The results are summarized in Table~\ref{tbl-eval-methods}. Surprisingly, the \textit{Naive} creator performs on par with, and often better than, the more elaborate off-the-shelf creators. This indicates that the harness can steer a model to evolve with high leverage---no parameter training required---but that the \emph{degree} of evolution is governed by the model's own intelligence rather than by the guidance encoded in the evolution method. With the model fixed, swapping the creator barely changes the size of the gain, and over-engineering it can even hurt.

\section{Conclusion}
\label{sec-conclusion}

We have presented \bench, an evolution-native benchmark and fully automated data pipeline for evaluating whether agents learn from prior experience and transfer that knowledge to new tasks. \bench is the first benchmark to evaluate agent self-evolution on GDP-related tasks, and the first to introduce \textbf{rule hybridization}, which makes generalization from training to test concrete and test-time gains attributable. Its fully automated pipeline further allows the benchmark to be regenerated quickly, providing a practical response to data contamination. We publicly release the pipeline, benchmark, and full evaluation results to support reproducible and economically grounded research on agent self-evolution.

\bibliographystyle{unsrt}
\bibliography{main}

\appendix
\clearpage
\section{Reproducibility and Artifact}
\label{sec-appendix-artifact}

The full \bench pipeline, the 240-task benchmark, the rule-based graders, and the released evaluation runs are publicly available. Each task group ships its shared environment, 5 training tasks, 5 held-out test tasks, and per-task evaluation scripts whose rubrics enumerate the weighted scoring points described in Section~\ref{sec-bench-grading}. The evaluation workspace is natural-language-driven: an orchestration agent opens a folder of Markdown skills and guides and launches an experiment from a natural-language request. Every skill-generation run and test-solving attempt uses a dedicated Docker container, staged directory, harness home, and trace path.

\section{Task Group Details}
\label{sec-appendix-tasks}

Table~\ref{tbl-data} lists all 24 released task groups. V1 contains four CRM
groups, four ERP groups, and four Finance groups; V2 adds four Healthcare groups,
four Legal groups, and four Data Analysis groups. Each group contains 5 training and
5 held-out test tasks, for a total of 240 tasks.

As a representative V1 example, the CRM lead-capture group (tg01) hides atomic rules for sponsor-status priority, blacklist exclusion, contact deduplication and normalization, and follow-up scheduling. Its training tasks each exercise a subset of these rules, and its test tasks recombine them. The per-task rubrics make every atomic rule independently checkable, enabling the traceable and reproducible scoring used throughout the paper.

\begin{table}[t]
\centering
\small
\begin{tabular}{clcc}
\toprule
\textbf{ID} & \textbf{Scenario focus} & \textbf{Macro domain} & \textbf{Train/Test} \\
\midrule
tg01 & Marketing lead capture            & CRM     & 5 / 5 \\
tg02 & B2B quote \& account response     & CRM     & 5 / 5 \\
tg03 & Service ticket resolution         & CRM     & 5 / 5 \\
tg04 & Retention \& churn analytics      & CRM     & 5 / 5 \\
tg05 & Finance expense control           & ERP     & 5 / 5 \\
tg06 & Procurement \& receiving control  & ERP     & 5 / 5 \\
tg07 & Inventory \& order fulfillment    & ERP     & 5 / 5 \\
tg12 & HR employee lifecycle             & ERP     & 5 / 5 \\
tg08 & Tax \& estate advisory            & Finance & 5 / 5 \\
tg09 & Operational modeling \& reporting & Finance & 5 / 5 \\
tg10 & Investment strategy \& risk       & Finance & 5 / 5 \\
tg11 & Branch credit risk \& lending     & Finance & 5 / 5 \\
tg13 & Patient intake \& transfer         & Healthcare & 5 / 5 \\
tg14 & Payer authorization \& appeals     & Healthcare & 5 / 5 \\
tg15 & EHR quality \& data governance     & Healthcare & 5 / 5 \\
tg16 & Clinical protocol decision support & Healthcare & 5 / 5 \\
tg17 & White-collar investigation review  & Legal   & 5 / 5 \\
tg18 & Court disposition \& financial entries & Legal & 5 / 5 \\
tg19 & Regulatory licensing \& compliance & Legal   & 5 / 5 \\
tg20 & M\&A contract review \& negotiation & Legal  & 5 / 5 \\
tg21 & Data cleaning \& quality pipelines & Data Analysis & 5 / 5 \\
tg22 & SQL analytics \& reconciliation    & Data Analysis & 5 / 5 \\
tg23 & Public-health statistical audit    & Data Analysis & 5 / 5 \\
tg24 & Engineering portfolio analytics    & Data Analysis & 5 / 5 \\
\midrule
\multicolumn{2}{l}{\textbf{Total: 24 task groups}} & 6 domains & \textbf{120 / 120} \\
\bottomrule
\end{tabular}
\caption{The released \bench benchmark: 240 tasks across 24 stateful task
groups in six domains. Each task group shares one business environment, 5
training tasks, and 5 held-out test tasks, so that the benchmark can measure
whether evolving from earlier tasks improves later work in the \emph{same}
environment. Every task is paired with a deterministic rule-based grader
(Section~\ref{sec-bench-grading}).}
\label{tbl-data}
\end{table}

\clearpage
\section{Evaluation Results Details}
\label{sec-appendix-full-results}

Table~\ref{tbl-full-leaderboard} gives the complete numerical counterpart of the
project website's leaderboard.  It retains all four supervision types and reports
the model--harness composition, reasoning setting, accuracy and uncertainty,
lift over \base, and the principal test-time efficiency measures.

\begin{table}[H]
\centering
\scriptsize
\setlength{\tabcolsep}{3.2pt}
\renewcommand{\arraystretch}{1.08}
\begin{tabular}{@{}ll l rrrrr@{}}
\toprule
& & & \multicolumn{2}{c}{\textbf{Accuracy}} & \multicolumn{3}{c}{\textbf{Test-time efficiency}} \\
\cmidrule(lr){4-5}\cmidrule(lr){6-8}
\textbf{Model (setting) + Harness} & \textbf{Supervision type} & \textbf{Evolution method} &
\textbf{ACC $\pm$ STD} & \textbf{Lift (pp)} & \textbf{USD/task} & \textbf{Rounds} & \textbf{Tokens (k)} \\
\midrule
\multirow{4}{*}{\makecell[l]{\textcolor[HTML]{3B5B92}{$\bullet$}\ \textbf{GPT-5.5} (xhigh)\\\hphantom{$\bullet$\ }Codex}}
& \base                    & --            & $49.37 \pm 5.51$ & --      & 1.29 & 14.96 & 767.9 \\
& \texttt{fewshot}         & Skill Creator & $64.51 \pm 6.31$ & $+15.14$ & 1.02 & 12.04 & 543.9 \\
& \texttt{self}            & Skill Creator & $55.80 \pm 7.63$ & $+6.42$  & 1.00 & 11.45 & 487.3 \\
& \texttt{reflect-3}       & Skill Creator & $57.82 \pm 7.38$ & $+8.45$  & 1.07 & 11.85 & 526.8 \\
\addlinespace[2pt]
\multirow{4}{*}{\makecell[l]{\textcolor[HTML]{845EA4}{$\bullet$}\ \textbf{Opus-4.8} (xhigh)\\\hphantom{$\bullet$\ }Claude Code}}
& \base                    & --            & $50.63 \pm 5.37$ & --      & 1.37 & 17.06 & 785.8 \\
& \texttt{fewshot}         & Skill Creator & $67.07 \pm 6.22$ & $+16.44$ & 1.36 & 14.46 & 819.1 \\
& \texttt{self}            & Skill Creator & $55.05 \pm 6.96$ & $+4.42$  & 1.49 & 15.57 & 964.0 \\
& \texttt{reflect-3}       & Skill Creator & $59.27 \pm 7.01$ & $+8.64$  & 1.43 & 15.58 & 938.2 \\
\addlinespace[2pt]
\multirow{4}{*}{\makecell[l]{\textcolor[HTML]{C85442}{$\bullet$}\ \textbf{GLM-5.2} (max)\\\hphantom{$\bullet$\ }Claude Code}}
& \base                    & --            & $46.12 \pm 5.82$ & --      & 0.50 & 22.94 & 1006.4 \\
& \texttt{fewshot}         & Skill Creator & $60.09 \pm 7.90$ & $+13.97$ & 0.53 & 22.63 & 1111.8 \\
& \texttt{self}            & Skill Creator & $50.96 \pm 8.32$ & $+4.84$  & 0.50 & 21.93 & 1037.8 \\
& \texttt{reflect-3}       & Skill Creator & $55.49 \pm 8.28$ & $+9.37$  & 0.47 & 20.65 & 952.1 \\
\addlinespace[2pt]
\multirow{4}{*}{\makecell[l]{\textcolor[HTML]{0F766E}{$\bullet$}\ \textbf{DS-V4-Pro-Preview} (max)\\\hphantom{$\bullet$\ }Claude Code}}
& \base                    & --            & $43.58 \pm 7.77$ & --     & 0.039 & 15.65 & 776.0 \\
& \texttt{fewshot}         & Skill Creator & $48.79 \pm 9.05$ & $+5.21$ & 0.040 & 14.73 & 764.6 \\
& \texttt{self}            & Skill Creator & $46.17 \pm 7.89$ & $+2.59$ & 0.040 & 14.29 & 777.1 \\
& \texttt{reflect-3}       & Skill Creator & $47.15 \pm 8.18$ & $+3.57$ & 0.037 & 12.68 & 635.3 \\
\bottomrule
\end{tabular}
\caption{Full task-group 001--024 leaderboard corresponding to the current project website. Each test task is run three times. ACC and STD preserve the task-group macro-averaging convention used by the released leaderboard; lift is the absolute percentage-point change from the matching \base row. Test-time cost and tokens exclude one-time skill generation.}
\label{tbl-full-leaderboard}
\end{table}

Table~\ref{tbl-evolution-cost} separately reports the one-time cost of producing
the skills used by the non-base supervision types.  These costs are excluded from the
test-time efficiency columns in Table~\ref{tbl-full-leaderboard}.

\begin{table}[H]
\centering
\small
\setlength{\tabcolsep}{7pt}
\renewcommand{\arraystretch}{1.22}
\begin{tabular}{@{}lrrrrrr@{}}
\toprule
& \multicolumn{2}{c}{\textbf{fewshot}} & \multicolumn{2}{c}{\textbf{self}} & \multicolumn{2}{c}{\textbf{reflect-3}} \\
\cmidrule(lr){2-3}\cmidrule(lr){4-5}\cmidrule(lr){6-7}
\textbf{Model + Harness} & \textbf{USD} & \textbf{M tok.} & \textbf{USD} & \textbf{M tok.} & \textbf{USD} & \textbf{M tok.} \\
\midrule
\textcolor[HTML]{3B5B92}{$\bullet$}\ \textbf{GPT-5.5} / Codex
  & 1.36 & 1.02 & 1.17 & 0.82 & 3.08 & 2.61 \\
\textcolor[HTML]{845EA4}{$\bullet$}\ \textbf{Opus-4.8} / Claude Code
  & 2.56 & 2.25 & 1.66 & 1.31 & 5.41 & 5.03 \\
\textcolor[HTML]{C85442}{$\bullet$}\ \textbf{GLM-5.2} / Claude Code
  & 0.45 & 1.11 & 0.38 & 0.95 & 2.16 & 6.04 \\
\textcolor[HTML]{0F766E}{$\bullet$}\ \textbf{DS-V4-Pro-Preview} / Claude Code
  & 0.03 & 0.52 & 0.02 & 0.38 & 0.10 & 4.70 \\
\bottomrule
\end{tabular}
\caption{One-time skill-generation overhead per task group. Each value is averaged first over three independent generation attempts and then over task groups 001--024. Costs are in USD and tokens are in millions. A generated skill is reused for the five held-out tasks in its group, so the amortized per-test-task overhead is one fifth of each reported value. \base has no skill-generation stage.}
\label{tbl-evolution-cost}
\end{table}

\paragraph{Machine-readable results.}
The released aggregates and pointers to the corresponding structured reports
are maintained in the public
\href{https://github.com/Prism-Shadow/GDPevo/blob/main/experiments/EXPERIMENT_BOARD.md}{experiment board};
the report directories contain the task-group-level metrics and generated skill
artifacts used to construct both tables.

\section{How Is the Fully Informed Oracle Ceiling Obtained?}
\label{sec-appendix-human}

We construct a \emph{fully informed oracle ceiling} by pairing the Codex
harness and GPT-5.5 at \texttt{xhigh} reasoning effort with one non-expert human
operator in each attempt. Multiple operators participate across the evaluation.
The condition estimates performance when the evidence needed to recover the
task-group rules is directly available, while the held-out task must still be
completed in the benchmark environment. Distributing attempts across operators
reduces dependence on the familiarity or workflow of any single person.

Each attempt is evaluated in a fresh context. The assigned operator and Codex
receive the complete packages of the five corresponding training tasks,
including their inputs, standard answers, notes, deterministic evaluators,
rubrics, supporting files, and declared dependencies. They additionally receive
the current test input and access to the running task environment. The current
test answer, test notes and evaluator, other test tasks, environment source,
task-group construction metadata, and records from previous attempts are not
included. The protocol therefore provides the available training-side evidence
without exposing the held-out solution, and it uses no separate training
execution or skill-generation stage.

The five training packages share the current task-group environment and expose
complementary parts of its hidden business logic. Standard answers demonstrate
the expected output form; notes, rubrics, and evaluators specify the constraints
that make those outputs valid; and supporting files and dependencies define the
records and tools available in the environment. The assigned operator can
inspect these materials while working with Codex on the held-out task. The
condition is therefore more informative than the main evaluation conditions,
but it does not reveal the target output or the grader used for the
current test.

The evaluation covers all 24 task groups, with five held-out tasks per group and
three independent attempts per task, for 360 attempts in total. Each attempt is
scored by the same deterministic grader used in the main evaluation. Although
the training-side evidence is available, each human--Codex pair must still find
the relevant records, determine which demonstrated rules apply to the current
instance, perform any required calculations, and produce a valid structured
output. Under this protocol, the human--Codex condition achieves an accuracy of
$91.6\%$.

\section{Breakdown and Diagnostic Views}
\label{sec-appendix-breakdowns}

\begin{figure}[H]
  \centering
  \includegraphics[width=0.59\textwidth]{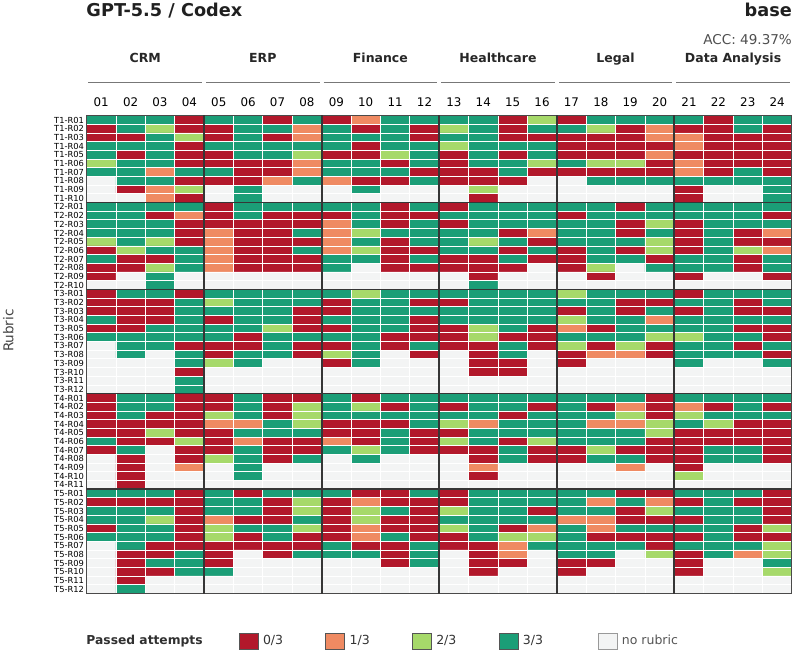}

  \vspace{0.25em}

  \includegraphics[width=0.59\textwidth]{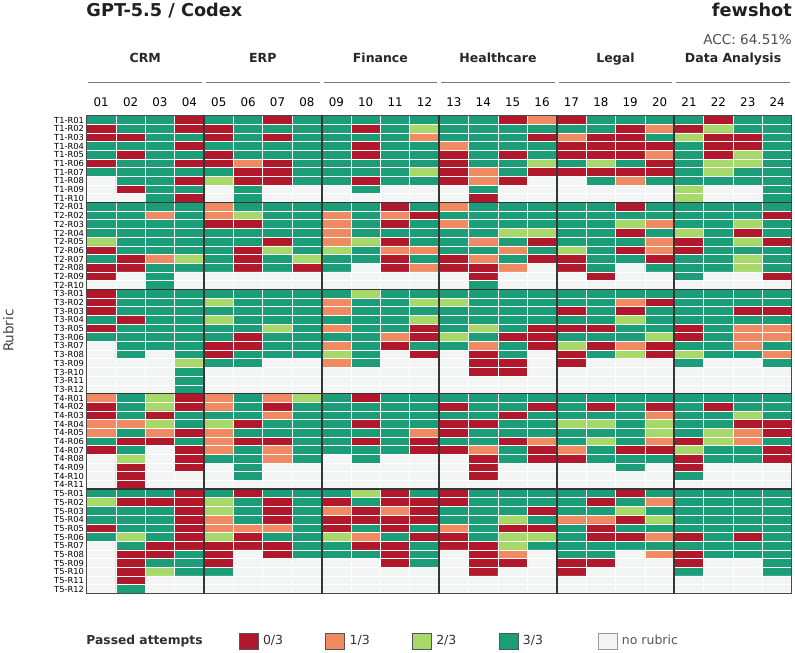}
  \caption{Rubric-level diagnostic views for GPT-5.5 with the Codex harness,
  comparing GPT-5.5 \base (top) and GPT-5.5 \fewshot (bottom) on task groups 001--024. Columns are task
  groups, arranged into six four-group domains; rows are test-local rubric
  indices and reset within each of the five held-out test tasks. Each colored
  cell reports how many of three independent attempts receive full credit on
  that binary rubric ($0/3$--$3/3$). Pale gray cells indicate that the
  corresponding test task has no rubric at that row, rather than a missing
  experimental result; all 24 task groups and all three attempts are present in
  both panels. Rubric weights are used to compute the reported accuracy, but color
  encodes the unweighted number of passing attempts so that individual rule
  reliability remains directly interpretable.}
  \label{fig:rubric-breakdown}
\end{figure}

Figure~\ref{fig:rubric-breakdown} exposes the rule-level structure behind the
aggregate evaluation results. The paired views hold the model and harness fixed
and show where \fewshot changes the consistency with which individual rubrics
are satisfied relative to \base.

\begin{figure}[H]
  \centering
  \includegraphics[width=\textwidth]{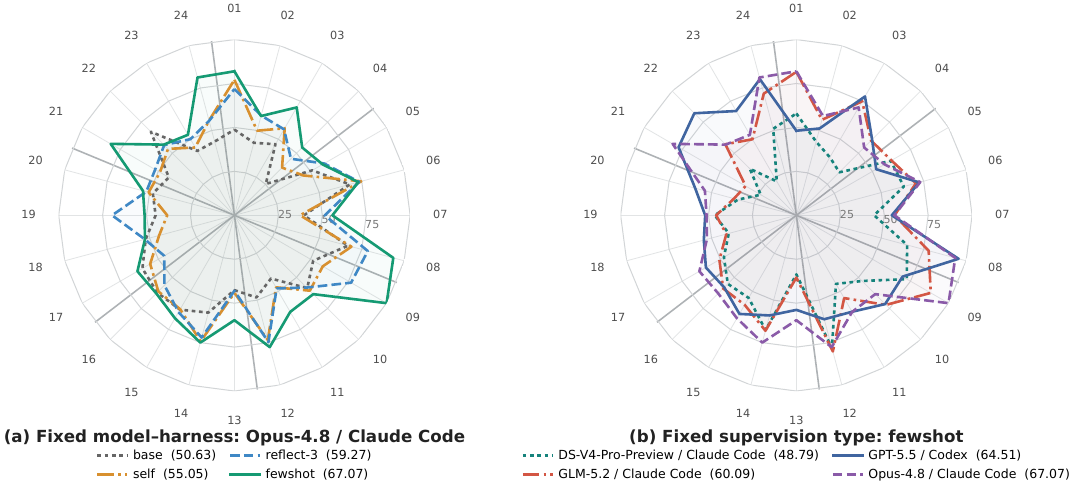}
  \caption{Task-group accuracy profiles on task groups 001--024. Panel (a)
  holds the model--harness configuration fixed at Opus-4.8 with Claude Code
  and compares the four supervision types. Panel (b) holds the supervision type
  fixed at \fewshot and compares the four model--harness configurations. Each
  spoke reports task-group accuracy on a shared 0--100 scale, averaged over
  three runs per test task; parenthetical legend
  values are macro averages over the 24 task groups. Spokes proceed clockwise
  through CRM (001--004), ERP (005--008), Finance (009--012), Healthcare
  (013--016), Legal (017--020), and Data Analysis (021--024).}
  \label{fig:task-group-radar}
\end{figure}

Figure~\ref{fig:task-group-radar} provides a complementary task-group view. For
Opus-4.8 with Claude Code, \fewshot raises macro accuracy from $50.63\%$ to
$67.07\%$ and outperforms \base on 23 of 24 task groups; \reflect and \self
improve 20 and 16 groups, respectively, with smaller macro gains. Holding
\fewshot fixed, Opus-4.8 attains the highest macro average, followed by GPT-5.5,
GLM-5.2, and DeepSeek-V4-Pro-Preview. The intersecting profiles show that these
aggregate advantages are broadly distributed but not uniform across task
groups.

\clearpage

\section{Evaluation Case Studies}
\label{sec-appendix-case-studies}

Aggregate accuracy does not reveal what a generated skill learns or why it
fails to transfer. We therefore examine four task groups using the generated
skills, training traces, held-out solver traces, and deterministic rubric
outputs. The cases cover two broad improvements and two forms of negative or
limited transfer. Because they use different model--harness configurations,
they are diagnostic examples rather than controlled comparisons between
models. Table~\ref{tab:appendix-case-overview} summarizes the score patterns.

\begin{table}[htbp]
  \centering
  \small
  \resizebox{\linewidth}{!}{%
    \begin{tabular}{@{}lllrrrr@{}}
      \toprule
      \textbf{Task group} & \textbf{Model + Harness} & \textbf{Observed transfer pattern}
      & \textbf{\base} & \textbf{\fewshot} & \textbf{\self} & \textbf{\texttt{reflect-3}} \\
      \midrule
      tg024 & GPT-5.5 / Codex & Conditional operational rules
      & 51.16 & 80.21 & 52.68 & 66.25 \\
      tg014 & GLM-5.2 / Claude Code & Cross-workflow rule recombination
      & 51.00 & 68.00 & 68.67 & 71.33 \\
      tg018 & DeepSeek-V4-Pro-Preview / Claude Code & Dropped scope conditions
      & 48.36 & 39.83 & 41.34 & 45.71 \\
      tg016 & GLM-5.2 / Claude Code & Limited hidden rule structure
      & 64.25 & 58.92 & 57.51 & 62.07 \\
      \bottomrule
    \end{tabular}%
  }
  \caption{Four diagnostic case studies. Values are held-out accuracy in
  percent, averaged over three runs per test task.
  The rows illustrate different transfer patterns and are not controlled
  model comparisons.}
  \label{tab:appendix-case-overview}
\end{table}

\subsection{TG024: Broad Transfer of Operational Decision Rules}
\label{subsec:appendix-good-case-024}

Task group 024 evaluates engineering-operations analysis with GPT-5.5 and the
Codex harness. The shared environment contains work items, teams, owners,
statuses, labels, SLAs, releases, milestones, blockers, and dependencies. The
five held-out tasks ask the agent to construct portfolio summaries, audit SLA
backlogs, assess release readiness, and prioritize overdue work. These tasks
require more than aggregation: the solver must identify authoritative fields,
exclude duplicate and cancelled records, classify work by its operational
purpose, and distinguish release gates from ordinary unfinished items.

The score pattern indicates broad positive transfer. GPT-5.5 \fewshot improves
held-out accuracy from the GPT-5.5 \base value of $51.16\%$ to $80.21\%$,
while GPT-5.5 \self and GPT-5.5 \reflect reach $52.68\%$ and $66.25\%$,
respectively. Its generated skill covers 33 of 35
training rules consistently and the remaining two partially. The improvement
also spans different workflows rather than a single repeated template.

Two tests expose what transfers. In \texttt{test\_001}, the solver must remove
cancelled and duplicate candidates before computing the portfolio mix. The
official population contains nine projects, classified as 0 NewFeature, 3
TechDebt, 4 Reliability, and 2 Security projects. The skill records that current
status, duplicate relations, and actual business purpose control inclusion and
classification; legacy categories are only references. An incorrect population
would change both the denominator and the investment recommendation.

In \texttt{test\_003}, nine work items are unfinished, but only four contain
high-impact blockers or incomplete critical dependencies. These four items are
hard release gates; the other five require monitoring but do not independently
prevent shipment. The learned rule therefore maps blocker evidence to a release
action rather than treating every unfinished item as equivalent. The resulting
decision is \texttt{NO\_SHIP} because unresolved hard gates remain.

This case illustrates a transferable skill as a sequence of conditional
decisions: validate the record, construct the population, classify by business
meaning, calculate only after the population is fixed, and act on risks that can
change the final decision. The skill also reduces search: GPT-5.5 \fewshot
averages approximately 15.53 tool calls per test, compared with about 30 for
GPT-5.5 \base.
The gain therefore comes with a more selective evidence-gathering procedure,
not additional exploration.

\subsection{TG014: Recombining Rules across Insurance Workflows}
\label{subsec:appendix-positive-case-014}

Task group 014 evaluates healthcare insurance operations with GLM-5.2 and
Claude Code. The tasks cover therapy authorization, drug-denial appeals,
workers-compensation claim recalculation, physician peer-to-peer review, and a
mixed work queue. The agent must both choose an operational action and provide
an audit trail that distinguishes supporting, missing, and excluded records.
All three supervision types improve over \base at $51.00\%$: \fewshot,
\self, and \reflect reach $68.00\%$, $68.67\%$, and $71.33\%$,
respectively.

One transferable rule concerns the granularity of missing evidence. In
\texttt{train\_002}, a broad packet item describes formulary-failure evidence,
but the gold output names the exact records that remain absent. In
\texttt{test\_002}, prior therapies are already documented for the Dupixent
appeal, so the insurance appeal no longer lacks medication evidence. Household
income proof remains missing only for the parallel manufacturer-assistance
workflow. A skill that repeats a broad packet label, or treats the assistance
gap as an appeal blocker, produces a plausible route but an incorrect audit
trail.

A second rule concerns evidentiary grounding. In \texttt{train\_004}, a
narrative note that PET imaging ``may be better'' is insufficient; the output
must cite structured records for each required PET-over-SPECT factor. In
\texttt{test\_004}, new peer-to-peer materials support overturning the denial
because they fill those factor-level requirements, not because they contain a
generally favorable narrative. Similarly, \texttt{train\_003} teaches that a
claim adjustment uses the current effective benchmark and computes each line as
the current allowed amount minus the amount already paid. The same operation is
reused in \texttt{test\_003} and inside the mixed queue of
\texttt{test\_005}.

The held-out tasks thus recombine rules learned from different training
workflows: name the exact missing record, keep parallel programs separate,
ground authorization decisions in structured factors, and recalculate payments
against the current benchmark. These rules are not tied to one patient, drug,
or claim line. They describe how the workflow should respond when the entities
change, which explains why all three generated skills outperform \base.

\subsection{TG018: Negative Transfer from Dropped Scope Conditions}
\label{subsec:appendix-overfit-case-018}

Task group 018 evaluates post-hearing court administration with
DeepSeek-V4-Pro-Preview and Claude Code. The solver reconciles case-system records, hearing
notes, clerk memoranda, and financial materials; determines whether a matter is
disposed, dismissed, continued, or pending; and then calculates charges,
payment schedules, docket actions, and forms. Jurisdiction, processing date,
petition type, and case status act as scope conditions that can change several
downstream outputs at once.

The generated skills do not transfer reliably. \base reaches $48.36\%$, while
\fewshot, \self, and \reflect reach $39.83\%$, $41.34\%$, and $45.71\%$.
For \fewshot, the gap is particularly diagnostic: training replay reaches
$73.88\%$, but held-out accuracy falls to $39.83\%$. The skill has learned
regularities that fit the training cases, but it often omits the conditions
under which those regularities apply.

Several learned statements turn local observations into global policies. A
source that is authoritative for identity is treated as authoritative for every
fact; fees absent from several training answers become a fixed ``never charge''
list; dates separated by roughly 30 days become a default deadline; and
reporting more conflicts is treated as safer. Each heuristic can match the
training jurisdictions while failing after the jurisdiction, policy date, or
case status changes.

The propagation is visible in \texttt{test\_004}. Case
\texttt{LC-25-0331} is continued and has no final order, so it cannot be closed
or charged in advance. Treating it as final changes the disposition, financial
entry, system action, and document together. The expected output contains
exactly seven material conflicts, whereas the \fewshot skill encourages the
solver to report nine. On this test, DeepSeek-V4-Pro-Preview \fewshot scores
$17.78\%$, while DeepSeek-V4-Pro-Preview \base scores $40.00\%$. The stored
skill is therefore not merely incomplete; it
actively directs the solver toward decisions that are invalid in the current
case.

This case separates rule content from rule scope. A useful court-administration
skill must state which source controls each question, which jurisdiction and
effective date select a fee schedule, and which case status permits closure.
Remembering the action without these conditions produces confident negative
transfer.

\subsection{TG016: Limited Transfer from Narrow Clinical Rules}
\label{subsec:appendix-limited-case-016}

Task group 016 evaluates clinical triage and care coordination with GLM-5.2 and
Claude Code. The solver retrieves current patient facts, reads a runtime
protocol, and fills a structured disposition containing risk, medication,
tests, follow-up time, and evidence identifiers. Unlike the preceding cases,
many decisive clinical thresholds are already visible in the runtime protocol.
The training tasks therefore expose fewer hidden rules that can add value at
test time. \base reaches $64.25\%$, while \fewshot, \self, and \reflect
reach $58.92\%$, $57.51\%$, and $62.07\%$.

Hypokalemia provides the clearest example. In \texttt{train\_003}, potassium is
3.2 without arrhythmic symptoms, so the routine oral-repletion branch schedules
follow-up at 08:00. In \texttt{test\_003}, the latest final potassium value is
2.8 and arrhythmia symptoms are present. The runtime protocol now selects
urgent escalation, with clinician notification, EKG, telemetry or emergency
department evaluation, and follow-up at 18:00. Reusing the learned 08:00 value
preserves the output but loses the branch condition that made it correct.

The respiratory workflow exposes a second limitation. A training answer teaches
the exact recommended-test set for one pneumonia scenario, which can help keep
the structured output precise. It does not determine the emergency disposition,
risk label, current-record selection, or evidence identifiers in a new case;
those decisions follow from the visible protocol and current patient records.
The learned rule is locally valid but covers only a small part of the held-out
scoring surface.

Negative transfer in this case therefore does not require a broadly incorrect
medical rule. It can arise when the task offers little hidden reusable structure
and the generated skill preserves a narrow value without its protocol branch.
When the current protocol already supplies the decisive rule, an additional
skill must remain subordinate to that evidence rather than override it.

\subsection{Cross-Case Synthesis}
\label{subsec:appendix-cross-case-discussion}

The four cases distinguish three requirements for reliable skill transfer.
First, the training tasks must expose reusable structure. TG024 and TG014
contain operational rules that recur after the entities and evidence change;
TG016 places more of its decisive structure directly in the runtime protocol.
Second, a generated skill must preserve scope. TG018 drops jurisdiction and
case-status conditions, while TG016 drops a clinical branch condition. Third,
the skill must connect evidence to the outputs it controls. The successful
cases specify which fields determine population membership, classification,
evidence placement, payment correction, or release action.

These patterns cannot be attributed solely to the task or model because the
cases use different model--harness configurations. They nevertheless suggest a
common audit for generated skills: each rule should identify its triggering
evidence, the output fields it changes, and the conditions under which it stops
applying. A skill that records all three behaves as a reusable decision rule; a
skill that records only a surface value or recurring action is more likely to
overfit or interfere with the current evidence.

\section{Lessons from Automated Self-Evolution Benchmark Generation}
\label{sec-appendix-generation-lessons}

Constructing the 24 task groups in \bench exposed recurring failure modes in automated benchmark generation. We summarize eight practical lessons under four themes: task construction, leakage control, grading and calibration, and evaluation integrity. Together, these lessons describe when an automatically generated task suite can support the claim that training experience improves held-out behavior.

\subsection{Designing Train--Test Relationships}

\paragraph{Train--test relationships must be designed from the outset.}
A self-evolution benchmark must construct a learning problem, not merely a collection of difficult tasks. In \bench, each task group contains five training tasks and five held-out test tasks in one shared business environment. The two sets share enterprise-specific rules, but differ in their entities, evidence, noise, and rule combinations. Rule hybridization creates this relationship by distributing atomic rules across training tasks and recombining them in test tasks.

Both extremes lead to misleading results. If training tasks directly demonstrate the test procedure, improvement may reflect template reuse. If training and test are unrelated, a lack of improvement does not show that the agent cannot evolve; it shows that the benchmark supplied no transferable experience. We therefore specify which test scoring points have training anchors, what knowledge transfers, and what changes at test time.

\paragraph{Generation and review should be independent.}
When one agent generates the environment, tasks, answers, graders, and review rationale in the same context, these artifacts may share the same assumptions and mistakes. A prompt may unintentionally echo the answer, or several rubric points may reward the same decision. \bench separates task-group design, environment construction, task construction, calibration, and final review. Independent reviewers inspect the completed group rather than accepting the builder's own assessment. This separation reduces builder--grader coupling and makes missing or inconsistent artifacts easier to detect.

\subsection{Controlling the Information Available during Evolution}

\paragraph{Isolation is more reliable than prompting.}
Asking an agent not to inspect hidden material is not a stable experimental control. For every attempt, we create a minimal workspace containing only the files allowed by the current stage and run the agent in an isolated container. Standard answers outside supervised training, notes, evaluators, unrelated tasks, and prior attempts are not mounted. The business environment runs separately and is accessible only through the allowed endpoints. Thus, agents with different file-exploration behavior still receive the same information.

\paragraph{Information access must match the declared supervision type.}
The four supervision types share the same task group, environment, grader, test protocol, and evolution method, but expose different training signals. \base receives no training experience; \fewshot receives the training inputs and gold answers; \self receives only the training inputs; and \reflect receives the training inputs together with a fixed budget of feedback on its own training attempts. These information boundaries must hold throughout the evaluation. If \self can access a training answer, or if \reflect can query the grader during testing, the configuration receives stronger supervision than declared and no longer measures the intended supervision type. A benchmark should therefore specify and audit the files, endpoints, feedback, and query budget available at every stage.

\subsection{Validating Grading and Calibration}

\paragraph{Rubric points should be deterministic, binary, and semantically distinct.}
A deterministic grader is reproducible, but reproducibility alone does not guarantee a meaningful score. Each scoring point in \bench represents a distinct business outcome and receives a raw weight from \(\{1,2,3\}\). It earns either its full normalized weight or zero; we do not award partial credit within a point. The task-factory prompt additionally requires 6--10 scoring points spanning at least four semantically distinct business outcomes and prohibits rewarding the same criterion, answer fact, or root decision more than once under different wording.

Together, these restrictions prevent a task-building agent from adjusting the score distribution through arbitrary subchecks, duplicate criteria, denominators, or tolerances. A submission may still receive a score between zero and one by passing a subset of independent points, but no individual point contributes a builder-chosen fraction. We validate the rubric by modifying one business outcome at a time and checking that the intended point changes while unrelated points remain unchanged.

\paragraph{Calibration must consider both difficulty and evolution gain.}
Reasonable \base accuracy is necessary but not sufficient. A task group may be difficult because information is missing or because its test tasks depend on rules that never appear in training. Conversely, near-perfect post-evolution accuracy may indicate that training and test are too similar. We therefore calibrate both initial difficulty and moderate, non-saturating improvement after training. When a group fails calibration, we revise its task design, evidence, or train--test relationship rather than changing rubric fractions to force the score into a target range. Calibration runs use fresh agent processes that do not inherit the builder's context.

\subsection{Keeping Evaluation Comparable and Auditable}

\paragraph{The evaluation agent must be capable and remain fixed.}
We adopt an \emph{AI-evaluates-AI} protocol in which a separate evaluation agent organizes the assessment of each evaluated agent. It stages files, launches containers, pairs generated skills with test attempts, invokes deterministic graders, handles failures, and aggregates results. The evaluation agent does not assign semantic scores itself, but its orchestration decisions can still affect the measured outcome. In particular, a less capable model in this role may mis-stage files, mismatch skills and attempts, fail to recover interrupted containers, or aggregate outputs incorrectly, causing evaluation failures unrelated to the capability being measured. We therefore use a sufficiently capable evaluation agent and keep its model, version, reasoning setting, prompt, tool permissions, retry policy, and aggregation rules fixed across comparable model--harness combinations. Changing the evaluated agent should not simultaneously change the agent that organizes the experiment.

\paragraph{Traces and costs should be preserved.}
Every evolution and test attempt receives a unique run identifier, an independent workspace, and a dedicated raw trace. A test process solves only one task, and each non-base test attempt uses the skill produced by the corresponding evolution attempt. Failed or contaminated attempts are retained for audit and replaced under the original protocol rather than silently scored as zero or dropped.

We also report skill-generation cost separately from test-time cost. The former measures the investment required to acquire a reusable skill; the latter measures the efficiency of applying it. Keeping the two stages separate makes it possible to determine whether a supervision type pays a larger one-time cost in exchange for repeated downstream savings.

In summary, automated self-evolution benchmark generation produces more than a task set. It also requires a designed train--test relationship, isolated information budgets, deterministic graders, calibration evidence, independent review, and attempt-level traces. These controls distinguish transferable improvement from answer leakage, template reuse, scoring artifacts, and execution failures.

\section{Use of Public Source Benchmarks}
\label{sec-appendix-sources}

We use public benchmarks only as sources for seed scenario discovery.  The source
pool spans professional deliverables and operating procedures, workplace
software and customer-support workflows, healthcare and legal operations, and
data-centric tasks.  Table~\ref{tbl-source-benchmarks} lists the 15 public
benchmarks represented in the released task groups and the type of seed each
source contributes.

For each scenario, the discovery stage selects a small number of source
examples and abstracts the work setting, artifacts, and operational constraints
that make the scenario professionally meaningful.  A scenario may combine
examples from multiple benchmarks.  These examples are not reused as evaluation
instances: the downstream pipeline constructs a new shared environment, five
training tasks, five held-out test tasks, reference answers, and deterministic
graders.  The source benchmarks therefore determine scenario coverage and
realism, while the evaluated instances and train--test relationships are newly
constructed for \bench.

\begin{table}[t]
\centering
\small
\setlength{\tabcolsep}{5pt}
\renewcommand{\arraystretch}{1.08}
\begin{tabularx}{\textwidth}{@{}>{\raggedright\arraybackslash}p{0.31\textwidth}X@{}}
\toprule
\textbf{Public source benchmark} & \textbf{Seed focus} \\
\midrule
GDPval~\citep{patwardhan2025gdpval}
  & Economically valuable professional deliverables \\
SOP-Bench~\citep{nandi2025sopbench}
  & Standard operating procedures and rule-governed workflows \\
SaaS-Bench~\citep{shi2026saasbench}
  & Multi-application business, healthcare, and software workflows \\
JobBench~\citep{li2026jobbench}
  & Occupation-specific professional tasks \\
Terminal-Bench~\citep{merrill2026terminalbench}
  & Command-line data preparation and analysis \\
$\tau^2$-Bench~\citep{barres2025tau2bench}
  & Tool-mediated customer-support workflows \\
CHI-Bench~\citep{chen2026chibench}
  & Healthcare administration and insurance workflows \\
MedAgentBench~\citep{jiang2025medagentbench}
  & Electronic-health-record operations \\
FHIR-AgentBench~\citep{lee2025fhiragentbench}
  & FHIR-based clinical record retrieval \\
Harvey LAB~\citep{harveylab2026}
  & Legal investigation and transactional workflows \\
SpreadsheetBench~\citep{ma2024spreadsheetbench}
  & Spreadsheet data preparation and quality control \\
InfiAgent-DABench~\citep{hu2024infiagentdabench}
  & Tabular data analysis and cleaning \\
BIRD-INTERACT~\citep{huo2025birdinteract}
  & Interactive analytical SQL tasks \\
LiveSQLBench~\citep{livesqlbench2025}
  & Stateful SQL and CRUD workflows \\
WorkBench~\citep{styles2024workbench}
  & Database-backed workplace analytics \\
\bottomrule
\end{tabularx}
\caption{Public benchmarks used during seed scenario discovery.  Source
examples motivate the work setting, artifacts, and constraints; \bench
constructs new environments, train--test tasks, reference answers, and graders.}
\label{tbl-source-benchmarks}
\end{table}


\end{document}